# Tracking Ethological Behaviors of Broiler Chickens Through Computer Vision (YoloV11) Using an Android Mobile Device

## Rastreamento de Comportamentos Etológicos de Frangos de Corte por Meio de Visão Computacional (YoloV11) Usando Dispositivo Mobile com Android

Ryan Davi Oliveira de Meneses[1]*, Glauber da Rocha Balthazar[1]*

**Abstract:** This study developed and evaluated a computer vision model for automated ethological behavior classification of broiler chickens in climate-controlled poultry houses. Using publicly available videos preprocessed and manually annotated into three classes (“sitting,” “standing,” and “feeding”), the YOLOv11 model was trained and deployed on an Android application via TFLite conversion. Exploratory analysis ensured dataset consistency and visual quality, while evaluation showed class-dependent performance: higher robustness for static postures (“sitting,” AP=0.531; precision=100%) and lower accuracy for dynamic behaviors (“feeding,” AP=0.292; recall=0.20). Tests with unseen images demonstrated significant improvements, achieving accuracies of 98.04% for “sitting,” 83.33% for “standing,” and 67.65% for “feeding.” The model showed effective generalization in real farm environments, supporting its potential for continuous behavioral monitoring, early welfare assessments, and decision-making in precision poultry farming. These findings establish a technical foundation for future mobile-based applications at commercial scale, reinforcing the role of computer vision in automated animal welfare evaluation.

**Resumo:** Este estudo desenvolveu e avaliou um modelo de visão computacional para identificação automatizada de comportamentos etológicos de frangos de corte em aviários climatizados. Utilizando v´ıdeos públicos pré-processados e rotulados em três classes (“sentado”, “de pé” e “consumindo”), o modelo YOLOv11 foi treinado e integrado em um aplicativo Android via TFLite. A análise exploratória dos dados assegurou consistência técnica e qualidade visual, enquanto os testes demonstraram desempenho diferenciado por classe: maior robustez na detecção de posturas estáticas (“sentado”, AP=0,531; precisão=100%) e menor acurácia em comportamentos dinâmicos (“consumindo”, AP=0,292; recall=0,20). Experimentos com imagens inéditas mostraram melhora significativa: acurácia de 98,04% para “sentado”, 83,33% para “de pé” e 67,65% para “consumindo”. O modelo evidenciou capacidade de generalização em ambientes reais, reforc¸ando seu potencial para monitoramento cont´ınuo, prevenção de problemas de bem-estar e suporte a` avicultura de precisão. Os resultados estabelecem base técnica para futuras aplicações móveis em escala comercial.

[1]*Departmento de Informa´tica, Instituto Federal de Sa˜o Paulo (IFSP), Brazil*
***Corresponding author**: meneses.ryan@aluno.ifsp.edu.br

## 1. Introduction

Computer vision, a branch of artificial intelligence focused on the automated analysis of images and videos, aims to extract relevant information, recognize patterns, and interpret complex scenarios [1]. Its applications include object classification, detection, and segmentation. In the poultry industry, this technology has been established as a strategic tool for monitoring and zootechnical management, enabling collective behavior analysis [2], physical condition assessment of birds [3], early detection of diseases and injuries [4], spatial occupancy measurement [5], feeding pattern monitoring [6], and body weight estimation [7].

In the ethological supervision of poultry, studies have employed video analysis to monitor locomotion, identify behavioral patterns, and detect anomalies that may indicate health or welfare concerns [2, 3, 7]. These approaches utilize recognition algorithms calibrated to distinguish individual birds and characterize their activities, enabling comparisons of trajectories and movement dynamics with previously established patterns for automated classification.

The development of such algorithms follows a structured workflow comprising key stages: video acquisition under diverse production conditions, expert-driven manual labeling of behaviors, application of computer vision techniques for bird detection, tracking, and classification, and, finally, real-time deployment through continuous video capture systems for automated identification and categorization of ethological patterns.

Recent research has strengthened the application of deep learning and computer vision in poultry ethology, particularly for automatic recognition of behavioral patterns in broiler chickens. Okinda et al. [8] presented one of the earliest comprehensive reviews on computer vision systems for poultry monitoring, outlining the transition from traditional image processing to deep neural networks. Subsequent studies have advanced this field using convolutional and transformer-based architectures to improve behavior classification accuracy. For instance, Guo et al. [9] and Nasiri et al. [10] proposed CNN-based models capable of distinguishing multiple behaviors such as feeding, resting, and standing, achieving accuracies above 85%. Elmessery et al. [11] applied YOLO frameworks to detect pathological and ethological phenomena in broilers through visible and thermal imagery, while Qi et al. [12] introduced a lightweight transformer model (FCBD-DETR) that enhanced both detection speed and precision. Similarly, Pearce et al. [13] demonstrated the feasibility of using machine learning to classify conventional and slow-growing broiler postures with over 90% accuracy. Broader approaches, such as the Ethoflow platform by Bernardes et al. [14] and trajectory-based clustering analyses in recent studies [15], further illustrate how vision-based systems are evolving toward continuous, large-scale behavioral monitoring. Collectively, these advances confirm that deep learning models, particularly YOLO variants and transformer-based architectures, provide an effective technical foundation for automatic ethological assessment and precision poultry farming applications.

This study hypothesized that computer vision could continuously monitor broiler chicken behavior to support health, welfare, and production. Accordingly, we developed, trained, and evaluated a model to automatically identify behavioral patterns in poultry facilities for implementation on Android devices. Using pre-processed and manually annotated public video datasets, the model was trained and then assessed with performance metrics in a simulated environment. The results provide a technical foundation for future automated systems that estimate indicators like body mass, stocking density, and feeding frequency.

## 2. Methods

### 2.1 Dataset

Videos of broiler chickens were sourced from public digital repositories (e.g., Kaggle, Zenodo) under open licenses permitting reuse. The dataset was pre-processed according to established literature protocols [16, 17], which involved removing duplicates and inadequate footage. Videos were excluded if they lacked birds, featured chicks younger than three weeks, or did not show relevant natural behaviors. The final dataset was restricted to footage from climate-controlled facilities.

### 2.2 Exploratory Data Analysis

An exploratory data analysis (EDA) approach was applied to characterize the dataset prior to training learning models. This process included organizing video sequences by facility type (conventional or climate-controlled), extracting technical metadata (resolution, frame rate, duration) for descriptive statistical analysis, selecting videos with a minimum visual quality of $\geq$360×480 pixels, and detecting technical anomalies in parameters such as contrast, color, and bit rate to ensure dataset consistency.

### 2.3 Training for Behavior Detection

Observed behaviors were categorized into three classes: standing, sitting, and consuming (feed or water). These categories guided the manual annotation of video frames, sampled every two seconds, to train a YOLOv11 model. The dataset was split into 70% for training and 30% for validation. Model performance was evaluated using standard metrics, including precision, recall, and mean Average Precision (mAP), to ensure robust detection and classification.

### 2.4 Implementation and Evaluation

After confirming effective classification, the trained model was deployed in an Android application using Java and the YOLOv4-TFLite API. The literature reports similar approaches that utilize Android-based mobile platforms for welfare assessment and broiler chicken growth monitoring through computer vision and thermal comfort analysis [18, 19]. During execution, the system recorded predictions as labeled sample frames, which allowed for a detailed analysis of correctly identified behaviors and false negatives. This method provided a quantitative evaluation of the model's accuracy and reliability in realistic conditions, demonstrating its suitability for automated behavior recognition.

## 3. Results and Discussion

An exploratory analysis characterized the video dataset used for model training. Video resolutions were predominantly

360×638, 360×640, and 720×1280 pixels, with frame rates between 27–30 FPS. Durations ranged from 8 to 31.57 seconds, providing sufficient temporal segments to capture complete behavioral sequences. This analysis confirmed the dataset's technical suitability for extracting ethological features.

From a compression perspective, all videos were encoded using the H.264/AVC standard with a *YUV420p* pixel format, ensuring broad compatibility. File sizes ranged from 0.5 MB to 11.57 MB, and all videos used progressive scan mode to eliminate artifacts. The color spaces were primarily bt709, and bit rates were sufficient to maintain quality for computational processing.

To ensure the integrity and reliability of the dataset used for YOLO model training and validation, a rigorous quality control pipeline was implemented to ensure dataset integrity. This process combined automated metadata analysis using *FFprobe* with manual visual inspection. We verified the homogeneity of video codecs (H.264/AVC), audio codecs (AAC), and scan types (progressive) to prevent processing errors. Visual checks confirmed that all videos contained broiler chickens exhibiting the three defined ethological behaviors. No critical anomalies were found, validating the dataset's robustness.

Model training performance was evaluated using multiple metrics: F1–Confidence Curves, Precision–Confidence Curves, Precision–Recall Curves, Recall–Confidence Curves, and confusion matrices. Figure 1 presents the relationship between F1-score values and classifier confidence thresholds for the three behavioral categories. The "sitting" class performed best, achieving a peak F1-score of approximately 0.55. The "standing" class showed intermediate and more variable performance with a peak F1-score of 0.43. The "consuming" class was the most challenging to detect, showing the lowest performance with a maximum F1-score of 0.36. Overall, the model achieved an optimal F1-score of 0.43 at a confidence threshold of 0.455.

Based on the Precision–Confidence (P-Curve) analysis shown in Figure 2, the pose estimation model demonstrated exceptional performance, achieving 1.00 precision for all keypoints at a 0.90 confidence threshold. The sitting class was the most robust, reaching approximately 0.95 precision at high confidence. The standing class showed intermediate performance, achieving 0.90 precision but with greater variability. The consuming class required higher confidence thresholds to achieve over 0.80 precision, reflecting its complexity. These results validate the model's suitability for automated monitoring in production environments.

The Precision–Recall (P–R) curve analysis in Figure 3 further demonstrates moderate overall model performance, with a mean Average Precision at an Intersection over Union threshold of 0.5 (mAP@0.5) of 0.417 across all classes. This result indicates general effectiveness in behavior detection, though significant variation is observed across individual classes. The sitting class achieved the highest performance (0.531), followed by standing (0.428) and consuming (0.292).

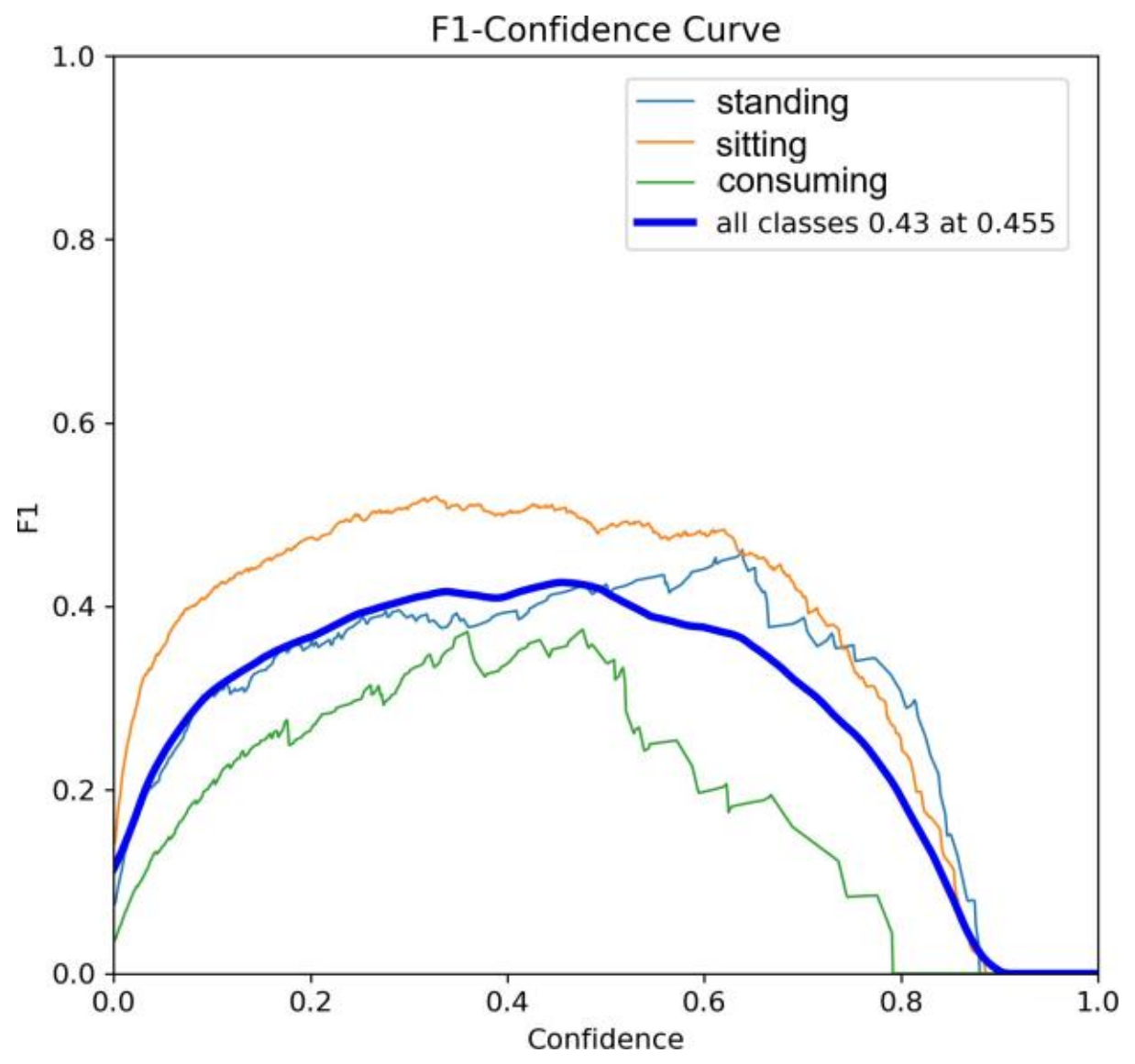

**Figure 1.** F1-Confidence Curve

The global mAP@0.5 = 0.417 reflects moderate detection capability, which, while promising for automated monitoring, highlights the need for further optimization, particularly to improve discrimination of complex feeding behaviors under commercial poultry production conditions.

Figure 4 illustrates the recall curve as a function of confidence thresholds for all evaluated classes. At low confidence thresholds, the model achieved high recall (0.89), indicating strong detection capability but with a higher likelihood of false positives. The sitting class (orange line) maintained the highest sensitivity at intermediate thresholds (recall values above 0.7). In contrast, the standing class (light blue line) had intermediate recall ( 0.6 at medium thresholds), and the consuming class (green line) showed the lowest recall, decreasing below 0.5 as confidence increased. These findings highlight that the model prioritizes static postures over dynamic behaviors and underscore the need to calibrate confidence thresholds to balance sensitivity and specificity.

The confusion matrix and normalized confusion matrix analyses (Figures 5 and 6) demonstrate strong but asymmetric performance in behavioral classification. The sitting class had high precision (0.88) and recall (0.61), with its primary confusion occurring with the background (219 false positives; 61% overlap). The consuming class was challenging, with low precision (0.51) and recall (0.20) due to significant confusion with both "standing" (0.40; 37 false negatives; 16% overlap) and background (0.24; 68 false positives; 19% interference) classes. The standing class had moderate accuracy (0.51) and low recall (0.47), suffering from high misclassification as background (48%) and confusion with "sitting" (26%), reflecting the difficulty in differentiating similar postures.

The model generalized effectively across various climate-controlled poultry facilities, robustly detecting sitting (cyan bounding boxes), standing (blue boxes), and consuming (red boxes) behaviors under different lighting, density, and layout conditions as shown in eight representative test images (clim-91 to clim-110, Figure 7). The high prevalence of sitting detections aligns with established broiler ethological patterns, confirming the biological validity of the results. The system successfully processed multiple, partially occluded individuals in dense populations, demonstrating its potential for integration into continuous monitoring systems for precision poultry farming.

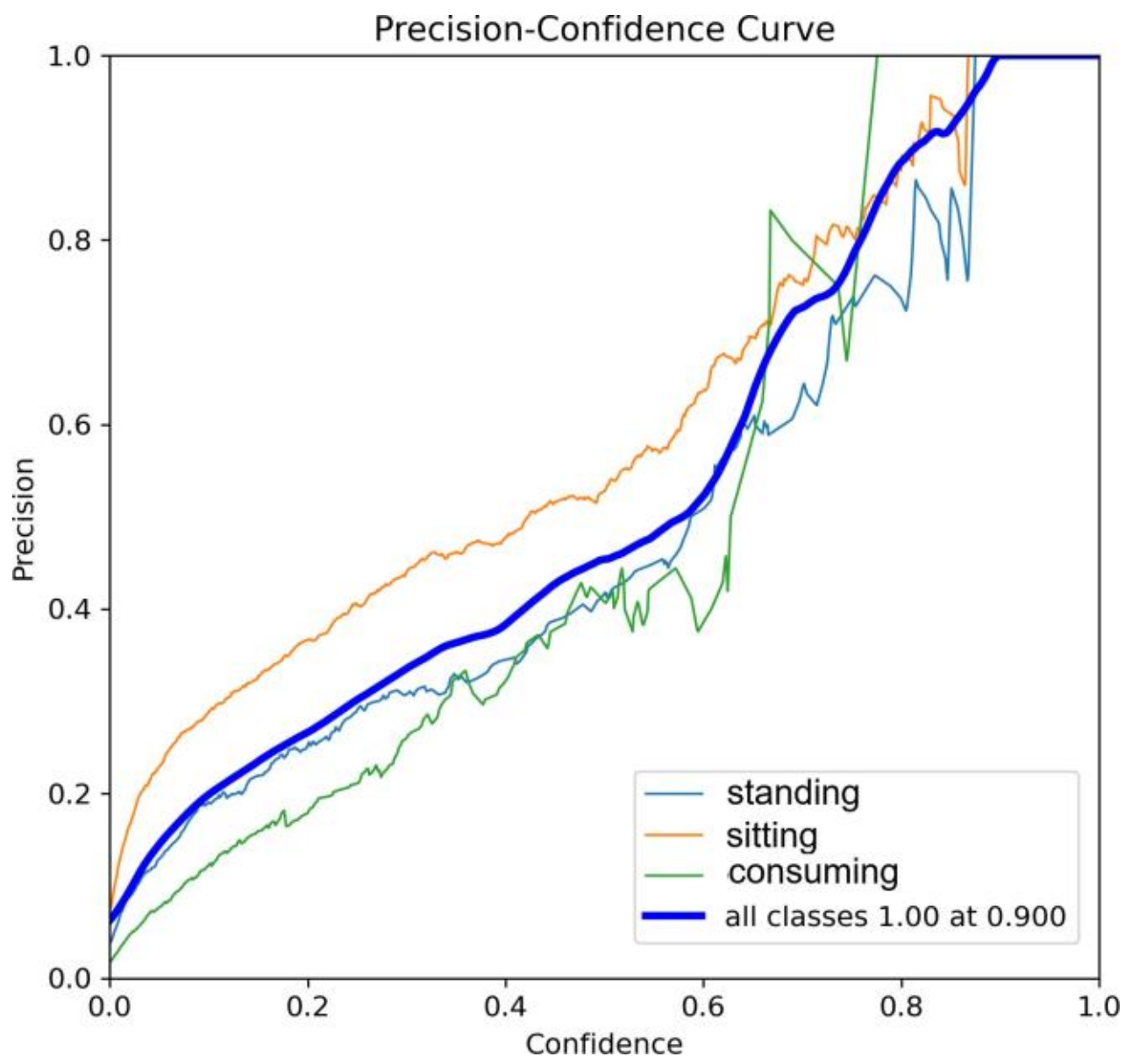

**Figure 2.** Precision confidence curve

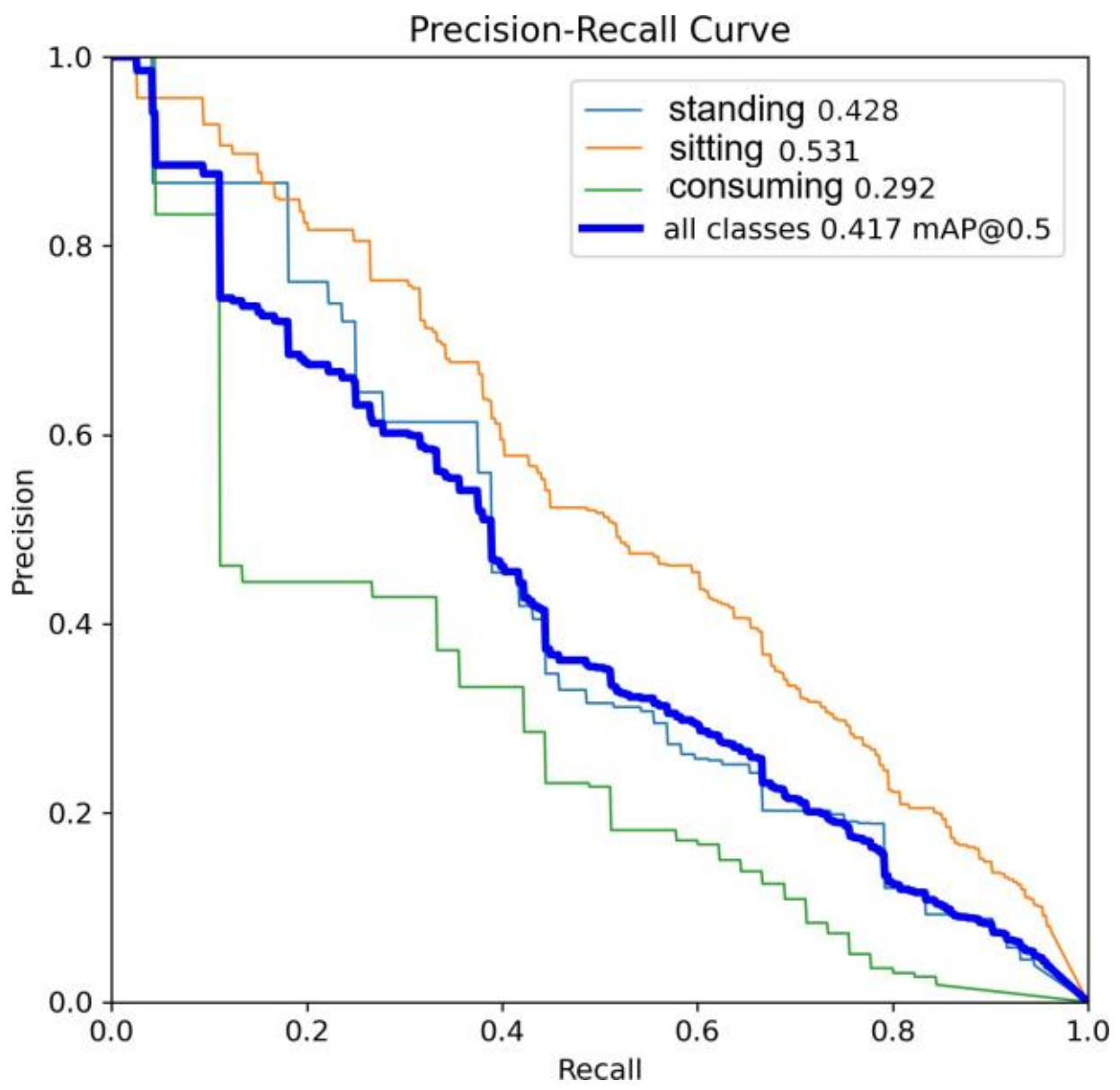

**Figure 3.** Precision recall curve

The overall detection performance (mAP@0.5 $\approx$ 0.417; AP: sitting = 0.531, standing = 0.428, consuming = 0.292) supports trends described in the literature [3], [4]: static postures achieve higher detection precision than dynamic, short-duration behaviors. The higher accuracy for sitting and lower precision for consuming reflect the visual complexity and postural variability of feeding behaviors, aligning with [2], which reported challenges in differentiating feeding and drinking activities in dense flocks, a scenario analogous to the climate-controlled facilities analyzed in this study.

A final experiment was conducted using the trained YOLOv11 model on an unpublished dataset of broiler chicken images captured under conditions identical to the training set (climate-controlled aviaries, $>$3 weeks of age). The model was deployed on a smartphone running Android 14 (Upside Down Cake), following conversion to TFLite format to optimize computational efficiency and memory consumption through quantization and compression, enabling deployment on mobile devices with native hardware accelerators (DSP/ GPU /NPU). This approach aligns with [19], which highlights the importance of lightweight architectures for mobile and embedded applications. The evaluation dataset comprised 102 images containing all ethological classes (standing, sitting, consuming). Images were displayed on a 22” monitor, and the smartphone’s camera captured them while running the application.

For the consuming class, the model achieved 67.65% accuracy, 76.19% precision, and 72.73% recall. The error analysis revealed that false negatives were the most critical issue, as the model missed approximately 27% of true instances, likely due to occlusions or postural variations. False positives were also significant, suggesting some confusion with other scene elements. These results demonstrate that while the model performs well on mobile hardware, optimizing for higher recall remains essential to ensure robust real-time ethological monitoring.

The results for the “standing” class (Figure 9) revealed 41 true positives (correct class detections), 5 false positives (detection errors where the class was not present), 44 true negatives (correctly identified absences), and 12 false negatives (detection failures when the class was present). These values indicate that the model achieved a global accuracy of 83.33%, demonstrating that more than four-fifths of the predictions made are correct. The detailed performance metrics revealed a precision of 89.13%, indicating that when the model detects the “standing” class, there is a very high probability of correctness. The recall of 77.36% shows that the model adequately identifies more than three-quarters of real class instances, while the specificity of 89.80% evidences a notable capability to correctly distinguish class absence. The main limitation was false negatives (22.64% of positive cases), suggesting detection failures due to postural variations or occlusion. In contrast, false positives were low (10.20%), indicating the model rarely confuses other objects with the “standing” class.

The results for the “sitting” class (Figure 10) revealed 97 true positives (correct class detections), 0 false positives

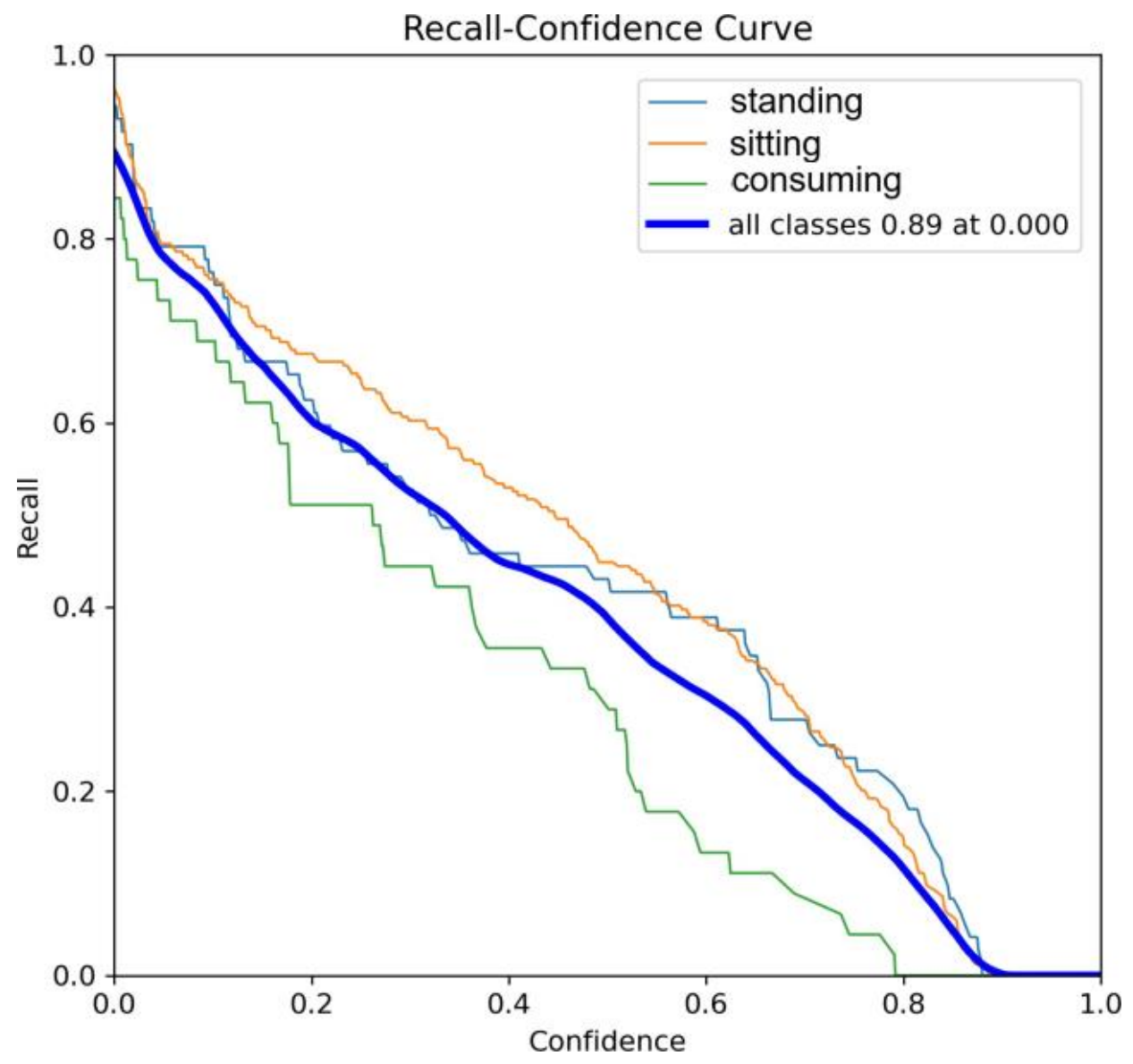

**Figure 4.** Recall confidence curve

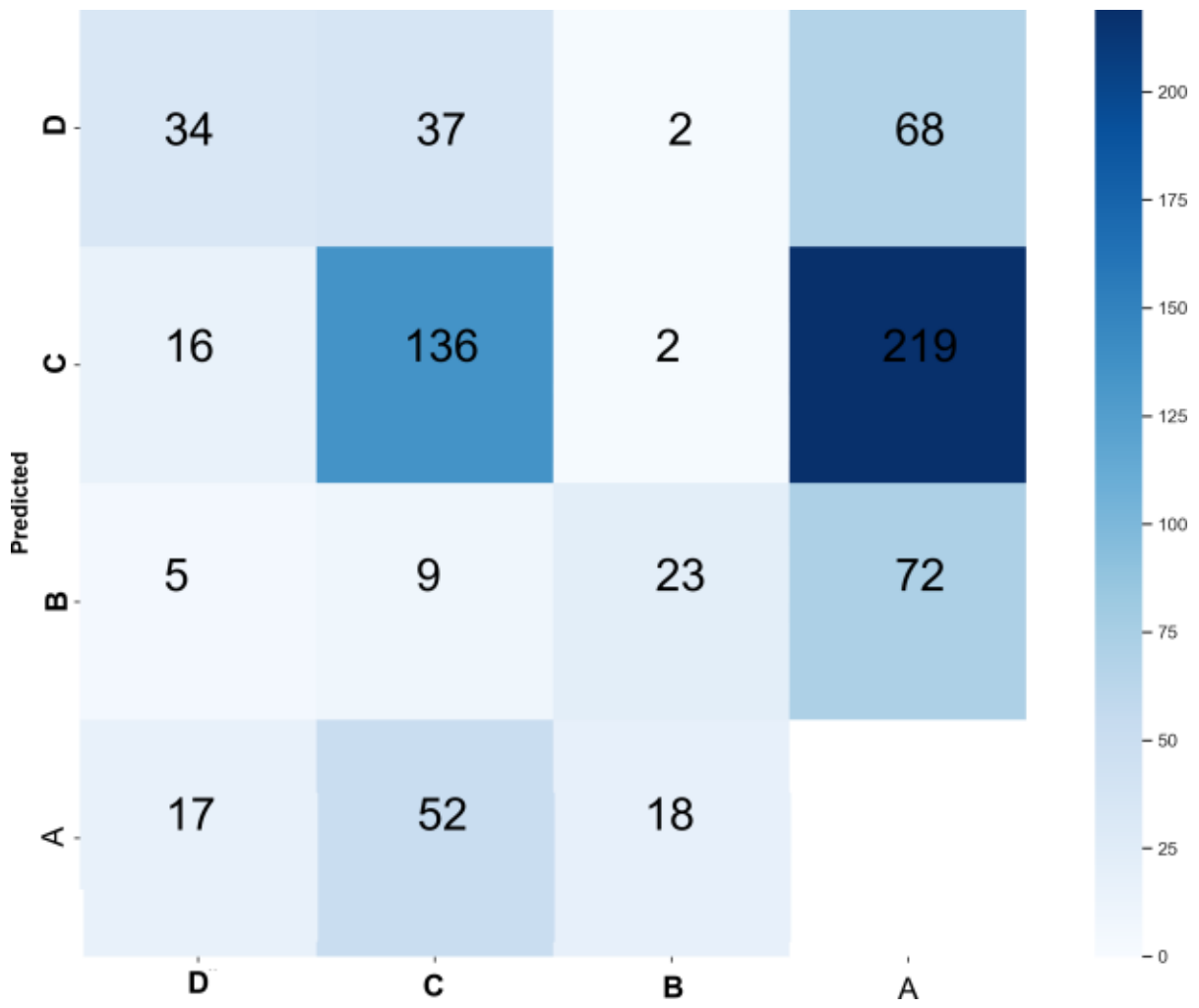

A: background; B: consuming; C: sitting; D: standing
**Figure 5.** Confusion Matrix

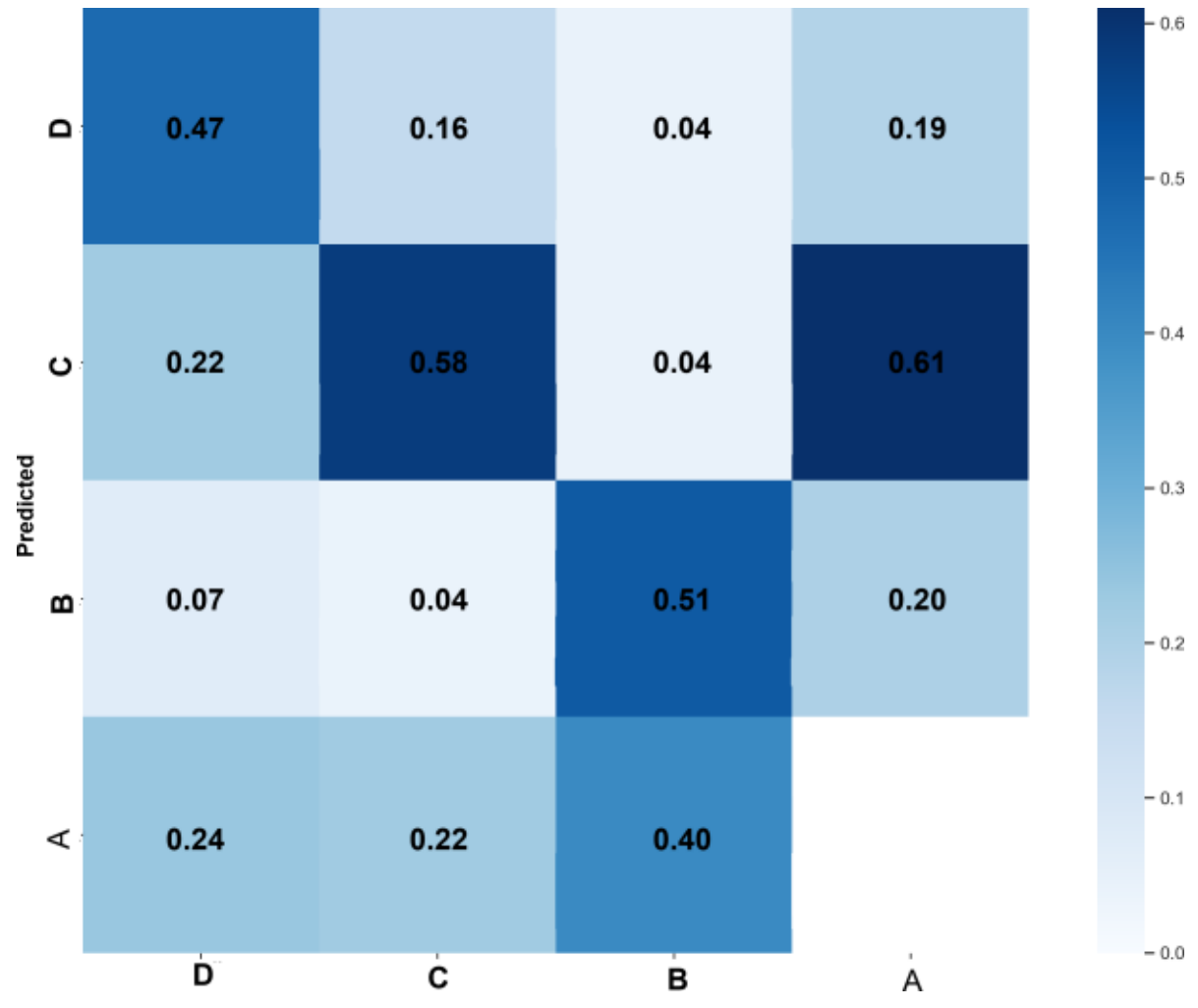

A: background; B: consuming; C: sitting; D: standing
**Figure 6.** Normalized confusion matrix

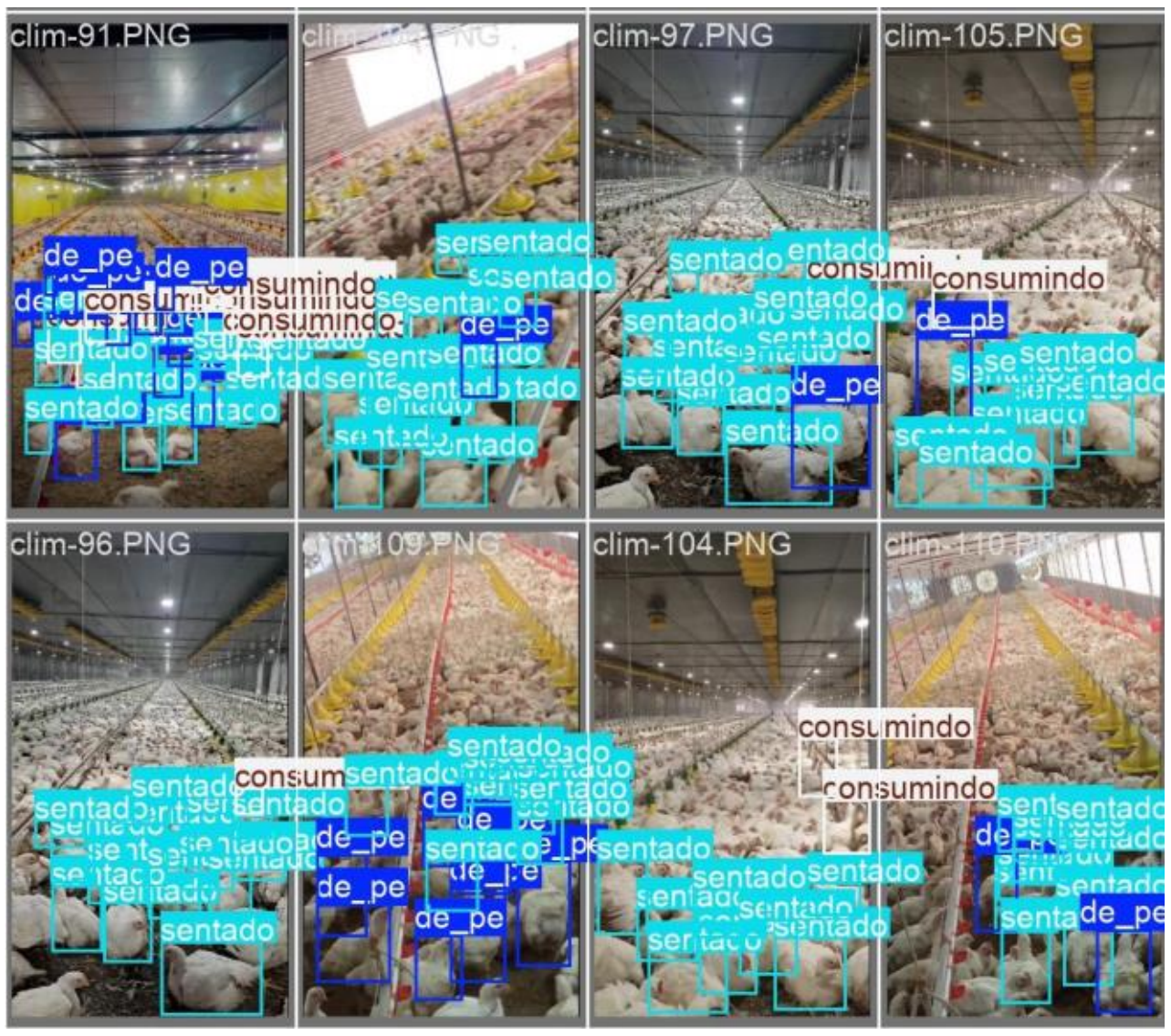

**Figure 7.** Model validation with different installations

(detection errors where the class was not present), 3 true negatives (correctly identified absences), and 2 false negatives (detection failures when the class was present). These values indicate that the model achieved a global accuracy of 98.04%, demonstrating that the majority of predictions made were correct. The detailed performance metrics revealed a perfect precision of 100%, indicating that when the model detects the "sitting" class, there are no false positive occurrences. The recall of 97.98% shows that the model adequately identifies almost all real class instances, while the specificity of 100% evidences an absolute capability to correctly distinguish class absence. The error analysis showed that false negatives constitute the only significant limitation, representing only 2.02% of real positive cases.

The asymmetric performance observed in the confusion matrix—marked by high precision for sitting and lower accuracy for consuming—reflects both technical and biological factors inherent to broiler behavior. Modern broilers exhibit prolonged inactivity as body weight increases, with resting postures (sitting) dominating their ethogram [13], [8]. Reduced mobility caused by leg problems and skeletal strain further decreases feeding-related movements, leading to partially reclined or static feeding postures that are visually similar to resting behaviors [4], [9]. Consequently, the model's misclassification between sitting and consuming may not solely indicate algorithmic limitations but rather mirror real behavioral overlap within intensive rearing systems. Similar misidentification trends were also reported by [10] and [12], where static

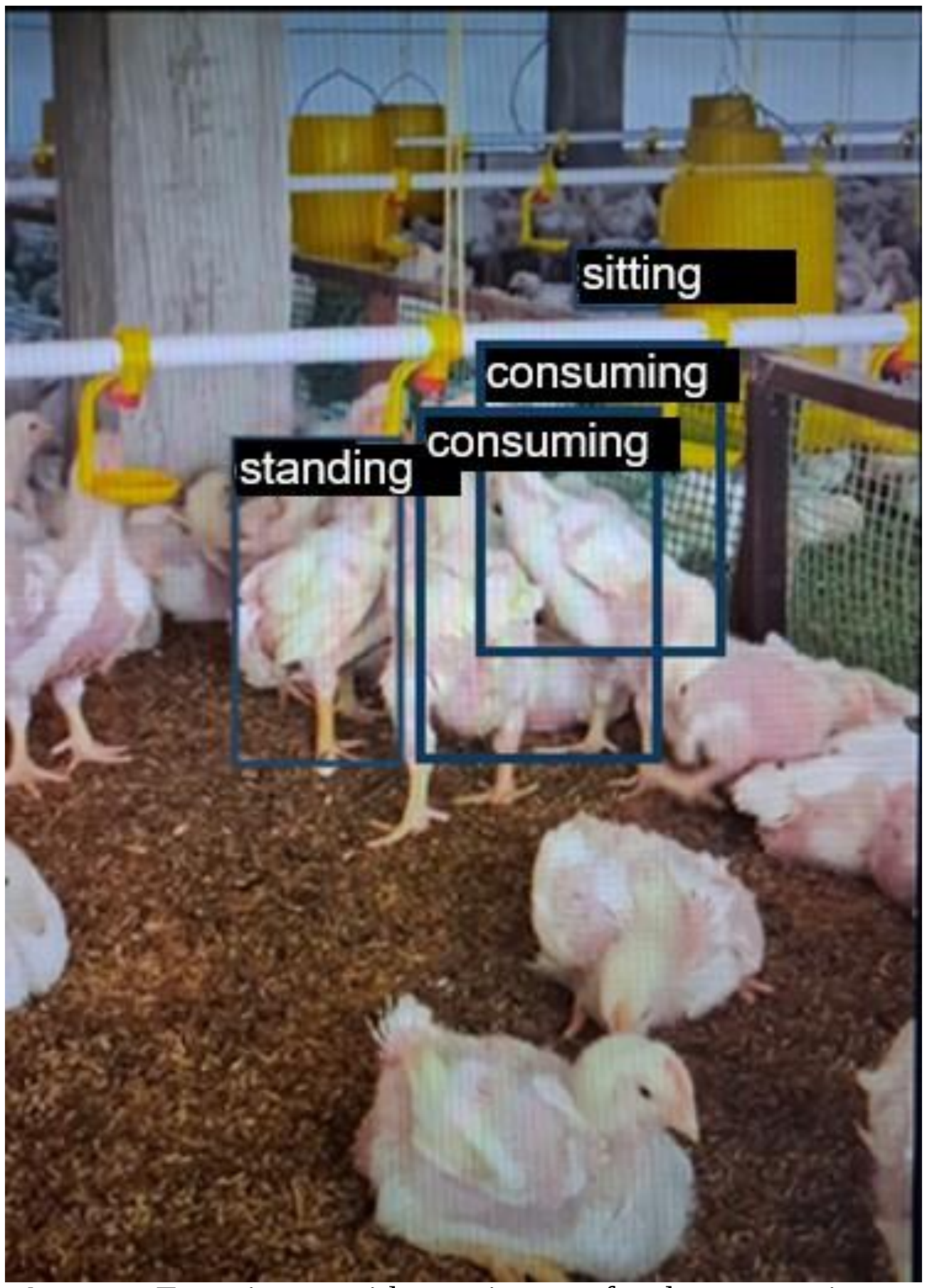

**Figure 8.** Experiment with new images for the consuming class

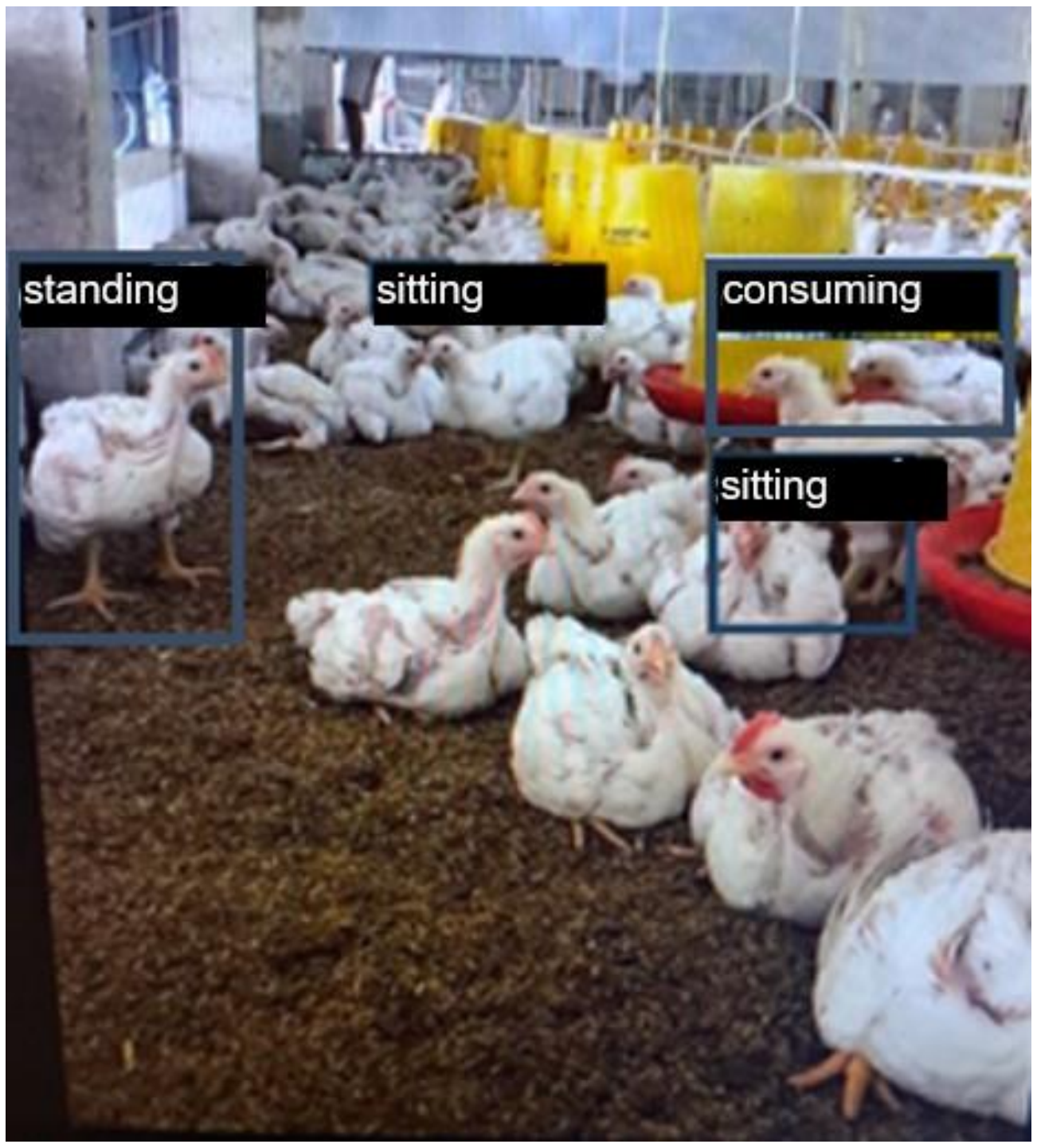

**Figure 9.** Experiment with new images for the standing class

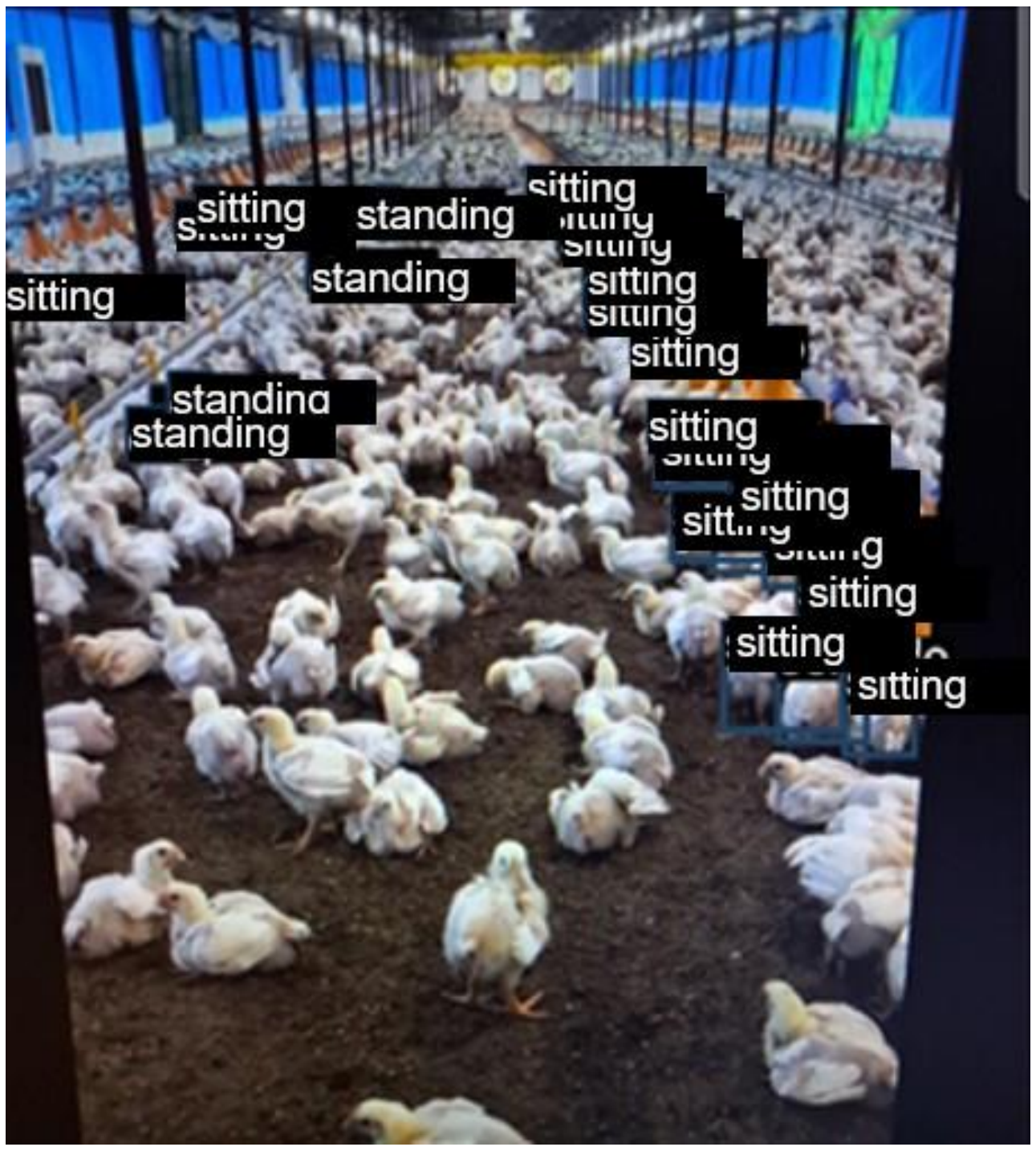

**Figure 10.** Experiment with new images for the sitting class

feeding behavior and low motion cues challenged classifier discrimination.

From a production standpoint, these findings underscore the necessity of interpreting automated behavioral monitoring within ethological and welfare contexts. Overestimation of resting behavior due to misclassified feeding may lead to biased welfare assessments or misguided management adjustments, such as ventilation or density changes. However, the model's robustness in detecting static postures provides a reliable basis for inactivity indices and early welfare diagnostics. As suggested by [11] and [12], integrating visual models with physiological or locomotion metrics (e.g., gait scoring, thermal data) could mitigate behavioral ambiguity and improve precision in commercial poultry monitoring. Ultimately, aligning computer vision outputs with biological interpretation enhances the model's applicability for precision livestock farming, supporting adaptive decision-making and proactive welfare management.

## 4. Technical Details

All model training and experimentation were performed on a workstation configured with an AMD Ryzen 5 5600G central processing unit (CPU) and an AMD Radeon RX 6600 graphics processing unit (GPU) with 8 GB of VRAM. The system was equipped with 16 GB of DDR4 RAM and operated on Windows 10. The training process utilized the YOLOv11M architecture, starting from pretrained model weights. Key hyperparameters for the object detection task included a batch size of 8, an input image size of 640×640 pixels, and a train-ing duration of 100 epochs. The total training process on this hardware configuration was completed in approximately four hours.

**Table 1.** Table of Grades

| Class | F1-Confidence | Precision confidence | Precision recall | Recall confidence | Confusion Matrix | Normalized confusion matrix |
|---|---|---|---|---|---|---|
| Sitting | 0.55 | 0.95 | 0.531 | 0.7 | 0.88 | 0.61 |
| Standing | 0.43 | 0.90 | 0.428 | ~0.6 | 0.51 | 0.47 |
| Consuming | 0.36 | 0.80 | 0.292 | 0.5 | 0.51 | 0.20 |

## 5. Conclusions

This study demonstrated that computer vision enables automated identification of broiler chicken behaviors in production environments, showing strong potential for continuous ethological monitoring. Exploratory analysis confirmed the dataset's consistency and suitability for model training.

Model evaluation revealed higher robustness in detecting static postures, particularly the sitting class (F1 $\approx$ 0.55; AP = 0.531), while dynamic behaviors such as consuming showed lower accuracy (AP = 0.292), mainly due to visual complexity and postural overlap. Overall performance (mAP@0.5 = 0.417) aligned with previous studies in animal ethology using computer vision. The disparity in performance, particularly the low accuracy for the consuming class, warrants deeper analysis as it directly impacts practical application.

Deployment on an Android smartphone improved classification results, with accuracies of 98.04% for sitting, 83.33% for standing, and 67.65% for consuming, confirming satisfactory generalization to unseen data. These results validate the hypothesis that deep learning models can effectively support automated behavioral monitoring in broilers and establish a foundation for scalable, mobile-based precision poultry farming applications.

Overall, the study validates the hypothesis that computer vision approaches can support automated ethological monitoring of broiler chickens, establishing a robust technical foundation for future precision poultry farming applications.

## Author Contributions

Ryan Davi Oliveira de Meneses contributed to programming the computer vision system, conducting YoloV11 training, and writing the document. Glauber da Rocha Balthazar contributed to formulating the work method, systematically and conceptually reviewing the work, and validating the results. All authors contributed to reviewing the work and approved the submitted version.